\documentclass[10pt,twocolumn,letterpaper]{article}

\usepackage{wacv}              

\usepackage{graphicx}
\usepackage{amsmath}
\usepackage{amssymb}
\usepackage{booktabs}
\usepackage{makecell}
\usepackage{cuted}
\usepackage{capt-of}
\usepackage{subcaption}
\usepackage[hypcap=false]{caption}
\usepackage{pifont}
\usepackage{xcolor}
\usepackage{color, colortbl}

\usepackage[accsupp]{axessibility}

\usepackage[pagebackref,breaklinks,colorlinks]{hyperref}

\usepackage[capitalize]{cleveref}
\crefname{section}{Sec.}{Secs.}
\Crefname{section}{Section}{Sections}
\Crefname{table}{Table}{Tables}
\crefname{table}{Tab.}{Tabs.}

\def\wacvPaperID{23} 
\def\confName{WACV}
\def\confYear{2024}

\begin{document}

\title{Task-Oriented Human-Object Interactions Generation with \\ Implicit Neural Representations}

\author{
Quanzhou Li$^1$ \qquad Jingbo Wang$^2$ \qquad Chen Change Loy$^1$ \qquad Bo Dai$^2$\\
$^1$ S-Lab, Nanyang Technological University \quad $^2$ Shanghai AI Laboratory\\
{\tt\small \{quanzhou001, ccloy\}@ntu.edu.sg \quad \{wangjingbo, daibo\}@pjlab.org.cn}
}
\maketitle

\begin{strip}\centering
\includegraphics[width=\textwidth]{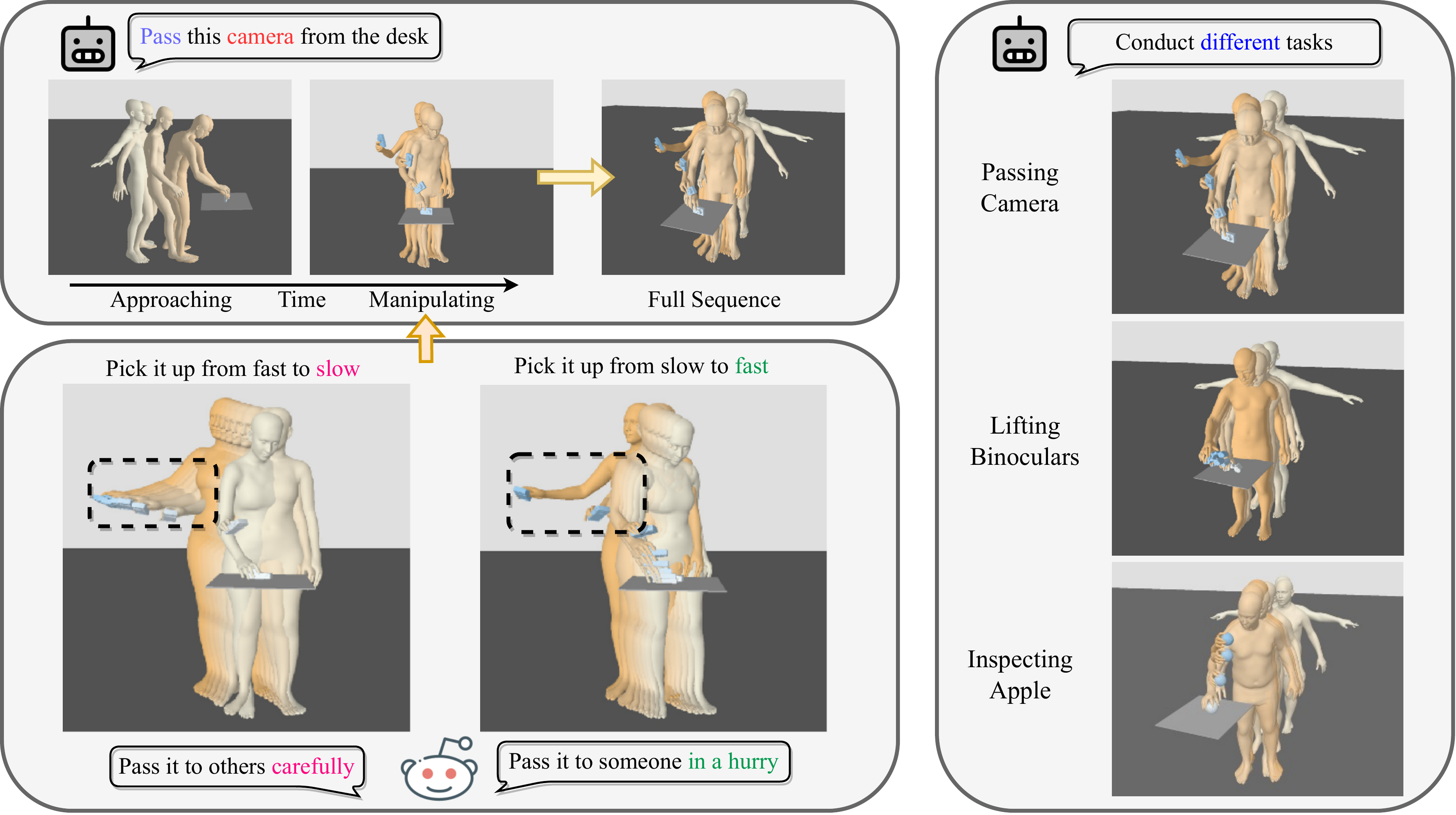}
\captionof{figure}{TOHO generates realistic and continuous task-oriented human-object interaction motions. To achieve a task, the human needs first to approach the object and then manipulate it. Unlike previous works which only synthesize the approaching motions or manipulation motions with known objects, TOHO generates complete task-oriented object manipulation sequences with unseen objects. In addition, as TOHO generates continuous sequences, we demonstrate that by inputting specific temporal coordinate vectors, TOHO can synthesize motions of various velocities and be upsampled to much higher framerates. On the right, we present three generated examples of digital humans with divergent shapes using novel objects to perform distinct tasks.
}\label{figure:intro}
\end{strip}

\begin{abstract}
\vspace{-0.2cm}
   Digital human motion synthesis is a vibrant research field with applications in movies, AR/VR, and video games. Whereas methods were proposed to generate natural and realistic human motions, most only focus on modeling humans and largely ignore object movements. Generating task-oriented human-object interaction motions in simulation is challenging. For different intents of using the objects, humans conduct various motions, which requires the human first to approach the objects and then make them move consistently with the human instead of staying still. Also, to deploy in downstream applications, the synthesized motions are desired to be flexible in length, providing options to personalize the predicted motions for various purposes. To this end, we propose TOHO: Task-Oriented Human-Object Interactions Generation with Implicit Neural Representations, which generates full human-object interaction motions to conduct specific tasks, given only the task type, the object, and a starting human status. TOHO generates human-object motions in four steps: 1) it first estimates the object's final position given the task intent; 2) it then generates keyframe poses grasping the objects; 3) after that, it infills the keyframes and generates continuous motions; 4) finally, it applies a compact closed-form object motion estimation to generate the object motion. Our method generates continuous motions that are parameterized only by the temporal coordinate, which allows for upsampling of the sequence to arbitrary frames and adjusting the motion speeds by designing the temporal coordinate vector. This work takes a step further toward general human-scene interaction simulation.
\end{abstract}

\section{Introduction}
\label{sec:intro}
 
Humans are in constant interactions with objects around them, and with different using intents, we conduct distinct motions with the objects. 
Generating such sequences in simulation is of great interest and value in various fields, from computer vision to robotics. Despite the tremendous progress in this topic in recent years, human-object interaction motion synthesis remains an under-studied problem. 
Previous works primarily focus on synthesizing human motions regardless of the objects or generating human interactions only with known objects already attached to the human hands. Moreover, existing works can only generate discrete frames and neglect actual motions' intrinsic diversity, making the synthesized motions hard to be manipulated in applications.

Synthesizing task-oriented human-object interaction motions is challenging due to various reasons. 
First, given an object with a task like lifting, the directions of moving the object are neither known nor unique and should be estimated by the given information. 
Second, a complete human-object interaction sequence should have the human approaching and grasping the object and making it move consistently with the human without floating around or staying still. Third, to deploy in applications that have various framerate requirements, generated motions should be flexible in length instead of having fixed framerates. 
Previous methods that are the most relevant to ours either only generate motions stopping at the grasping points, ignoring object motions and using intents \cite{taheri2021goal, wu2022saga}, or only generate object manipulation sequences with objects already attached to the human hands \cite{ghosh2023imos}, and neither can produce continuous results. Currently, synthesizing complete continuous task-oriented human-object interaction motions remains an unsolved problem.

Inspired by recent advances in human motion synthesis and implicit neural representations, we propose TOHO, a novel task-oriented human-object interaction motion generation framework. 
A natural human-object motion requires the human first to approach the object, grasp it, and then use it to accomplish the task, sequentially. TOHO addresses each part of the motion synthesis process, taking only the task type, the object at its initial position, and the initial human status as inputs. It generates natural and realistic human motions conducting a task with an unseen object while allowing the generated results to be of arbitrary lengths. 

We generate complete human-object motions with TOHO in four steps. 1) First, we design an object parameters sampler that estimates the object's translation and orientation offsets from its initial position with the task type and human shape, giving the final object position and orientation. 2) Similar to \cite{taheri2021goal, wu2022saga}, we use a goal pose generation network to generate grasping poses given the object information. 3) Then, we propose an INR-based motion inbetweening network that generates continuous motions to infill the missing in-between part of two frames. 4) Finally, we present a fast and compact object motion estimation algorithm that outputs realistic and natural object motions based on the generated human motions. With the algorithm, we synthesize perceptually realistic human-object manipulations with objects moving consistently with the human. 

With the four steps, our framework generates complete task-oriented human-object interaction motions. Unlike autoregressive and CNN-based methods, our framework generates motions that are continuous and parameterized only by the temporal coordinate. The continuity reflects what motions in the real world are like and allows for velocity adjustments and upsampling of sequences to arbitrary frames. Also, at inference time, as a frame is not conditioned on previous frames, all frames can be inferred parallelly with the same model. In Table \ref{table:checkbox}, we compare our problem setting with previous methods.

We summarize our contributions as 1) we present a unified framework to generate complete human-object interaction motions with intents; 
2) our framework generates continuous motions parameterized only by the temporal coordinates, exploring the intrinsic diversity of actual motions; 3) our generated results are flexible in length and speed, allowing upsampling and velocity adjustments in downstream applications.

\section{Related Work}
\label{sec:related_works}

\newcommand{\cmark}{\textcolor{green!80!black}{\ding{51}}}
\newcommand{\xmark}{\textcolor{red}{\ding{55}}}
\definecolor{Gray}{gray}{0.95}

\begin{table}[t]
\center   
\caption{
Overview of our problem setting compared with previous methods. Our method is the only unified framework that generates complete and continuous intent-driven human-object manipulation motions with unseen objects.
}
 \scalebox{0.72}{ 
 \begin{tabular}{lccccc}\toprule 
 & \multicolumn{5}{c}{ Motion Synthesis} \\
 				  \cmidrule(lr){2-6} 
 & \shortstack{Reaching\\Object} &\shortstack{Object\\Manipulation} &\shortstack{Unseen\\Object}  &\shortstack{Intent-\\Driven} &\shortstack{Continuous\\Motion}\\
 \addlinespace[2pt] \midrule
\rowcolor{Gray}
GOAL\cite{taheri2021goal} & \cmark &  \xmark &\cmark &  \xmark &  \xmark \\
SAGA \cite{wu2022saga}& \cmark  &  \xmark& \cmark & \xmark&  \xmark \\  
\rowcolor{Gray}
IMoS \cite{ghosh2023imos}& \xmark & \cmark & \xmark & \cmark &  \xmark \\
TOHO (ours)& \cmark & \cmark & \cmark & \cmark &  \cmark \\
\bottomrule 
\end{tabular}
}
\label{table:checkbox}  
\end{table}

\noindent\textbf{Hand Grasp Synthesis.} 
Synthesizing realistic hand grasps is a challenging problem, and with the advancement of deep learning, many works have been proposed to approach the task, including \cite{brahmbhatt2019contactgrasp, jiang2021graspTTA, kar2020grasp, GRAB:2020, turpin2022graspd, grady2021contactopt}. Taheri et al. \cite{GRAB:2020} propose a conditional variational autoencoder (cVAE) \cite{sohn2015cvae} based network to predict coarse MANO \cite{romero2017mano} parameters and then apply a refinement step to optimize them. Jiang et al. \cite{jiang2021graspTTA} present a method to estimate the hand pose together with a contact map, which is then used to refine the grasp. Going beyond the synthesis of static object grasps, some previous studies \cite{zhang2021manipnet, christen2022dgrasp} also explore the generation of hand grasp motions. Christen et al. \cite{christen2022dgrasp} suggest using reinforcement learning to synthesize physically plausible grasping motions. Given the object and wrists trajectories, Zhang et al. \cite{zhang2021manipnet} propose an autoregressive model to synthesize the hand-object manipulation motions.

While these works have well-addressed hand grasp synthesis, they only focus on the hands in isolation from the body. Our work differs from the earlier works in that we propose to generate whole-body human-object interactions, which lie in a higher parameter dimension and require consistency of all body parts.

\noindent\textbf{Human Motion Synthesis.} Human motion synthesis has become a prevailing research topic in recent years and has drawn attention from both computer vision and computer graphics 
\cite{martinez2017recurrent, gopalakrishan2019rnn, martinez2017rnn, kaufmann2020convolutional, li2021taskgeneric, cai2020unified, mao2020history, li2021choreographer, petrovich2021actor, tang2018longterm}.
Kaufmann et al. \cite{kaufmann2020convolutional} propose considering motion sequences as images and utilizing convolutional neural networks to inpaint the missing parts. Some studies \cite{li2021taskgeneric, cai2020unified} suggest the use of VAEs to generate stochastic human motions. Some more recent works \cite{mao2020history, li2021choreographer, petrovich2021actor, tang2018longterm} make use of the attention \cite{vaswani2017attention} mechanism to model human motions. Even though these works have made significant progress in human motion synthesis, they only focus on modeling humans regardless of the scenes and objects around them.

There are some existing works proposed to tackle human motion synthesis involving the 3D scenes, including 
\cite{cao2020longterm, harvey2020robust, li2019affordance, makansi2019overcoming, rempe2021humor, sadeghian2019sophie, starke2019neural,  wang2021synthesizing, zhang2021priors, wang2022sceneaware, hassan2021stochastic, corona2020context, taheri2021goal, wu2022saga}. 
Wang et al. \cite{wang2021synthesizing} propose a hierarchical motion synthesis framework, first synthesizing several sub-goal poses in 3D scenes, then infilling the whole motion sequence, and finally refining the motion sequence with an optimization scheme. Hassan et al. \cite{hassan2021stochastic} suggest first estimating the object's goal pose and contact to interact and a trajectory for the human to approach and then generating the human motion with an autoregressive module.

The works that are the most relevant to ours are \cite{taheri2021goal, wu2022saga, ghosh2023imos}. GOAL \cite{taheri2021goal} suggests first estimating a grasping pose and then synthesizing the motion of the human from its initial position to the goal with an autoregressive model. Similarly, SAGA \cite{wu2022saga} first predicts a grasping pose and uses a CNN-based cVAE to infill the motion. While both works generate realistic motions, they only model the human approaching the objects and stopping at the 'touching' points, ignoring human-object interactions. Ghosh et al. \cite{ghosh2023imos} develop the IMoS method to generate human-object interactions assuming a grasp on a known object is already established. IMoS employs an autoregressive prediction to generate 15 frames for one motion clip and linearly interpolate them into 30 frames. Unlike the abovementioned methods, our framework synthesizes complete intent-driven human-object interactions with unseen objects, and the generated sequence can be upsampled to arbitrary frames.

\noindent\textbf{Implicit Neural Representations.} Implicit neural representations have gained considerable attention recently with the success of SIREN \cite{sitzmann2020inr}, and NeRF \cite{mildenhall2020nerf}. The critical insight of INR is that a complex signal can be represented by a function of spatial or temporal coordinates at its corresponding position, and high-frequency details can be well preserved through this mapping. INR has demonstrated its efficacy on multiple tasks, including image synthesis \cite{inr_gan, anokhin2021image}, video generation \cite{yu2022digan}, time-varying 3D geometries \cite{niemeyer2019occupancy}, and dynamic scenes \cite{pumarola2021dnerf, li2022neural3d}. Recently, He et al. \cite{he2022nemf} propose a task-agnostic INR-based representation to interpret human motion as a function of time and conduct tasks through per-sequence optimization. Inspired by these works, we propose an INR-based generative model to infill the motion between two arbitrary human poses and positions.

\section{Method}

\label{sec:method}

\begin{figure*}
    \centering
    \includegraphics[width=\textwidth]{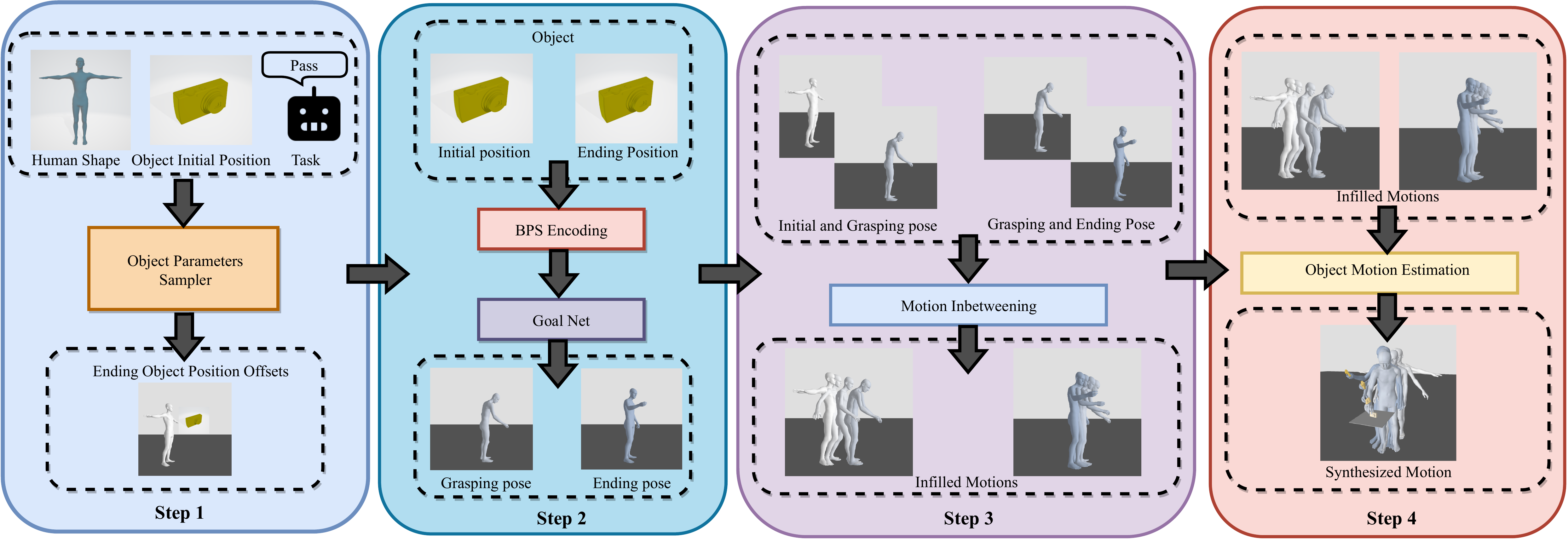}
    \caption{Overview of TOHO. We formulate the generation of object manipulation motions as a motion-infilling problem that consists of four steps. 1) With the shape parameters of the human and the task type, we estimate the object's final position using our object parameters sampler. 2) The goal net then generates human poses grasping the object at its initial and final positions. 3) Our motion inbetweening model subsequently generates continuous motions to infill the missing frames between the keyframe human poses. 4) Finally, our object motion estimation algorithm outputs a stable and consistent object motion based on the human motion in real-time.}
    \label{fig:pipeline}
\end{figure*}

\noindent\textbf{Preliminaries. 1) 3D human representation.} We use the SMPL-X model \cite{pav2019smplx}, which models the human body with hand and face details. SMPL-X takes human shape, $\beta$, pose, $\theta$, and body global translation, $t$, as inputs and generates a mesh with 10,475 vertices. In this work, we predict the 6D continuous pose rotation vector \cite{zhou2019continuity} $\theta \in \mathbb{R}^{55\times 6}$ and the global translation $t \in \mathbb{R}^{3}$. The shape parameter $\beta \in \mathbb{R}^{10}$ is constant for a fixed body shape. \textbf{2) Object shape representation.} We use the Basis Point Set (BPS) \cite{prokudin2019bps} distances to represent the object shapes. Following \cite{GRAB:2020}, we randomly sample 1024 vertices from $[-0.15, 0.15]^3$ as the basis point set to calculate distances.

\subsection{Overview}
The overview of our method is shown in Fig. \ref{fig:pipeline}. Our goal is to synthesize continuous human-object motions conducting specific tasks with unseen objects. Given three inputs, namely: 1) a task type, 2) an object shape with its starting translation and orientation, and 3) a human pose with its shape at its initial position, our method generates a continuous motion sequence allowing the virtual human to grasp the object and conduct the task with it. The framework generates motions in four steps. We introduce each of the steps in the following sections.

\subsection{Object Parameters Sampler} 
Previous works \cite{taheri2021goal, wu2022saga} propose to use a pose prediction network to estimate the human pose at the grasping point based on the object's shape and position. However, the object's final position is not conveniently given in object manipulation. Also, with different types of tasks, the object could be taken to very diverse positions. \cite{ghosh2023imos} synthesizes object manipulation motions by predicting frames autoregressively. However, it assumes a grasp is already established and only synthesizes 15-frame motions into the future with known objects. To synthesize long-term human-object motion sequences with high fidelity, we first generate keyframes of a complete manipulation task and infill the in-between frames, which makes the whole sequence bounded by the keyframes and would not deviate. To this end, we formulate the task of generating complete human-object manipulation motions as a motion inbetweening problem. To generate the keyframes, we first propose a task-conditioned object position estimator.
Our object parameters sampler is a cVAE, and in the  training stage it takes
\begin{equation}
X_s=[a_{\rm one}, \beta, t_o^{\rm init}, r_o^{\rm init}, t_o^{\rm off}, t_o^{\rm off} ]
\end{equation}
as inputs, where $a_{\rm one}$ is the $a$-th column of an identity matrix, i.e., a one-hot vector, $a \in \mathbb{R}$ is the task label, $\beta \in \mathbb{R}^{10}$ is the human shape, $ t_o^{\rm init} \in \mathbb{R}^3$ and $r_o^{\rm init} \in \mathbb{R}^6$ are the initial translation and orientation of the object, and $t_o^{\rm off} \in \mathbb{R}^3$ and $r_o^{\rm off} \in \mathbb{R}^6$ are the translation and orientation offsets of the object’s ending position from its initial position. The object position estimator encodes the input to a latent space with a dimensionality of 16 and is conditioned on $a_{\rm one}$, $\beta$, $ t_o^{\rm init}$, and $ r_o^{\rm init}$. In inference, the model uses a sampled latent code  $z_s \sim \mathcal{N}(\mu_s, \sigma_s^2)$, where $\mu_s \in \mathbb{R}^{16}$ and $\sigma_s \in \mathbb{R}^{16}$, and the conditioned parameters to decode the ending object translation offset $\hat{t}_o^{\rm off}$ and orientation offset $\hat{r}_o^{\rm off}$. 
The loss to training the sampler is defined as 
\begin{equation}
\mathcal{L}_s=\lambda_t||\hat{t}_o^f-t_o^f||_2 + \lambda_r||\hat{r}_o^f-r_o^f||_2  + \lambda_{KL}\mathcal{L}_{KL} \label{1},
\end{equation}
where $\mathcal{L}_{KL}$ is the Kullback-Leibler divergence loss. We set the loss coefficients empirically to balance the different terms.

\subsection{Goal Net} 
Given the object's position at the initial and final positions, we design a goal net that utilizes the object information to synthesize keyframe grasping poses. Our goal net is a cVAE conditioned on the object following \cite{taheri2021goal, wu2022saga }. During training, the goal net inputs the whole-body grasp with the objects’ shape and translation and reconstructs the poses and hand-object distances given the objects’ shape and location. The input to the encoder is:
\begin{equation}
X=[\theta, t, \beta, v, d_{b\to o}, h, t_o, b_o, a] \label{goal_net_input},
\end{equation}
where $\theta$, $t$, and $\beta$ are the human’s pose, global translation, and shape parameters, respectively, $v \in \mathbb{R}^{400\times 3}$ is the 3D coordinates of 400 sampled vertices on the human’s body surface, $h \in \mathbb{R}^3$ denotes the head orientation, $t_o \in \mathbb{R}$ is the object translation, $b_o \in \mathbb{R}^{1024}$ is the BPS representation of the object shape, $a \in \mathbb{R}$ is the task label, and $d_{b\to o} \in \mathbb{R}^{400\times 3}$ is the offset vectors from the sampled body vertices to the closest object vertices. 
The decoder predicts the SMPL-X parameters $\hat{\theta}$, $\hat{t}$, the head orientation $\hat{h}$, and a right-hand offset vector $\hat{d}_{r\to o} \in \mathbb{R}^{99\times 3}$ which is a subset of the 400 body vertices to object offsets, given a sampled latent code and conditioned on $\beta$, $b_o$, $t_o$, and $a$. We use the same loss in \cite{taheri2021goal} for training and follow its post-optimization scheme to refine the generated poses.

\subsection{Motion Inbetweening} 
Existing works of feed-forward motion inbetweening can only generate discrete motion clips. However, the framerate requirements could be diverse for real-world applications in AR/VR and gaming. Previous works need to either re-train the model or apply complex linear interpolations to synthesize a motion sequence of a different length \cite{wu2022saga}, or have to run a per-sequence optimization which prohibits real-time deployment \cite{he2022nemf}. To this end, we propose a motion inbetweening method that generates continuous motions that are parameterized only by the temporal coordinates to infill two given frames. The continuity of the model allows the generated motions to be upsampled to arbitrary frames and sampled non-uniformly to give motions of diverse speeds, all by inputting different temporal coordinates to the same inference model.

Given two frames, our motion inbetweening module aims to predict the human poses and root translations in frames between them. Following \cite{wu2022saga}, we find that first computing an interpolated root trajectory of the first and last frames and then predicting the translation offsets of each frame results in smoother results compared to directly predicting the translations themselves. To infill two frames, we view a motion clip as a motion image with the human pose parameters at frame $i$, $\theta_i \in \mathbb{R}^{55 \times 6}$, $i\in [1,...T]$, flattened and concatenated with its corresponding root translation $t_i \in \mathbb{R}^3$ as a column vector. For training, we define $T=64$, while in inference, $T$ can be any integer value. Thus, one motion clip is represented as a $333 \times 64$ motion image.
We design our motion inbetweening model as ${\rm F}(\tau;\phi _{\rm INR})=(\theta, t^{\rm off})$, where $\tau \in [0, 1]$ is the temporal coordinate, $\theta$ and $t^{\rm off}$ are the pose and the translation offset from the interpolated trajectory at time $\tau$, respectively. For training, we have each $\tau_i=i/64$ correspond to the $i$-th column of the motion image. The motion inbetweening net is a hypernetwork-based model: firstly, it takes poses of the two endpoint frames, $\theta_1$ and $\theta_T$, and their translation distance, $d_{1\to T}= t_T - t_1$, as input. Then, it generates the weights $\phi_{\rm INR}$ for an INR block ${\rm F}_{\phi_{\rm INR}}$. The INR takes the temporal coordinate vector $\tau_{1:T}$, $\tau_i \in [0, 1]$ as input and generates the complete motion image:
\begin{equation}
M=[\hat{\theta}_{1:T}, \hat{t}^{\rm off}_{1:T}], \label{3}
\end{equation}
where $\hat{\theta}_{1:T}$ and $\hat{t^{\rm off}}_{1:T}$ are reconstructed SMPL-X pose and translation offset sequences. We apply Factorized Multiplicative Modulation (FMM) \cite{inr_gan} to parameterize the weights of our INR to reduce the number of model parameters.

With the inferenced SMPL-X parameters, we then calculate 99 surface marker locations $\hat{v}^M_{1:T}\in \mathbb{R}^{99\times 3}$ and eight foot-ground contact labels $\hat{C}_{fg} \in \{0, 1\}^8$. This aids the model in utilizing information from the 3D space and assists in circumventing the skating problem. The loss to train the motion inbetweening model is defined as:
\begin{multline}
\mathcal{L}_M =\lambda_{\theta}\mathcal{L}_{\theta} + \lambda_{t}\mathcal{L}_{t}
 + \lambda_{v}\mathcal{L}_{v} + \lambda_{C}\mathcal{L}_{C},    \label{4}
\end{multline}
where $\mathcal{L}_{\theta}=\sum_{n=1}^T ||\theta_n - \hat{\theta}_n||_2$, $\mathcal{L}_{t}=\sum_{n=1}^T ||t_n - \hat{t}_n||_1$, $\mathcal{L}_{v}=\sum_{n=1}^T ||v^M _n - \hat{v}^M_n||_1$, $\lambda_{C}\mathcal{L}_{C}=\mathcal{L}_{bce}(C_{fg}, \hat{C}_{fg})$. The hat denotes the reconstructed values.

In inference, we replace the first and last frames of the sequence with the input two frames. Unlike \cite{wu2022saga}, which requires extensive post-optimization to get smooth motion sequences, we only add a lightweight post-processing step: to get a smoother result at the connection points, we interpolate the first and last five frames with the input frames. The motion inbetweening net is implemented with MLPs with skip connections. More details are given in the supplementary material.

\begin{figure*}
    \centering
    \includegraphics[width=\linewidth]{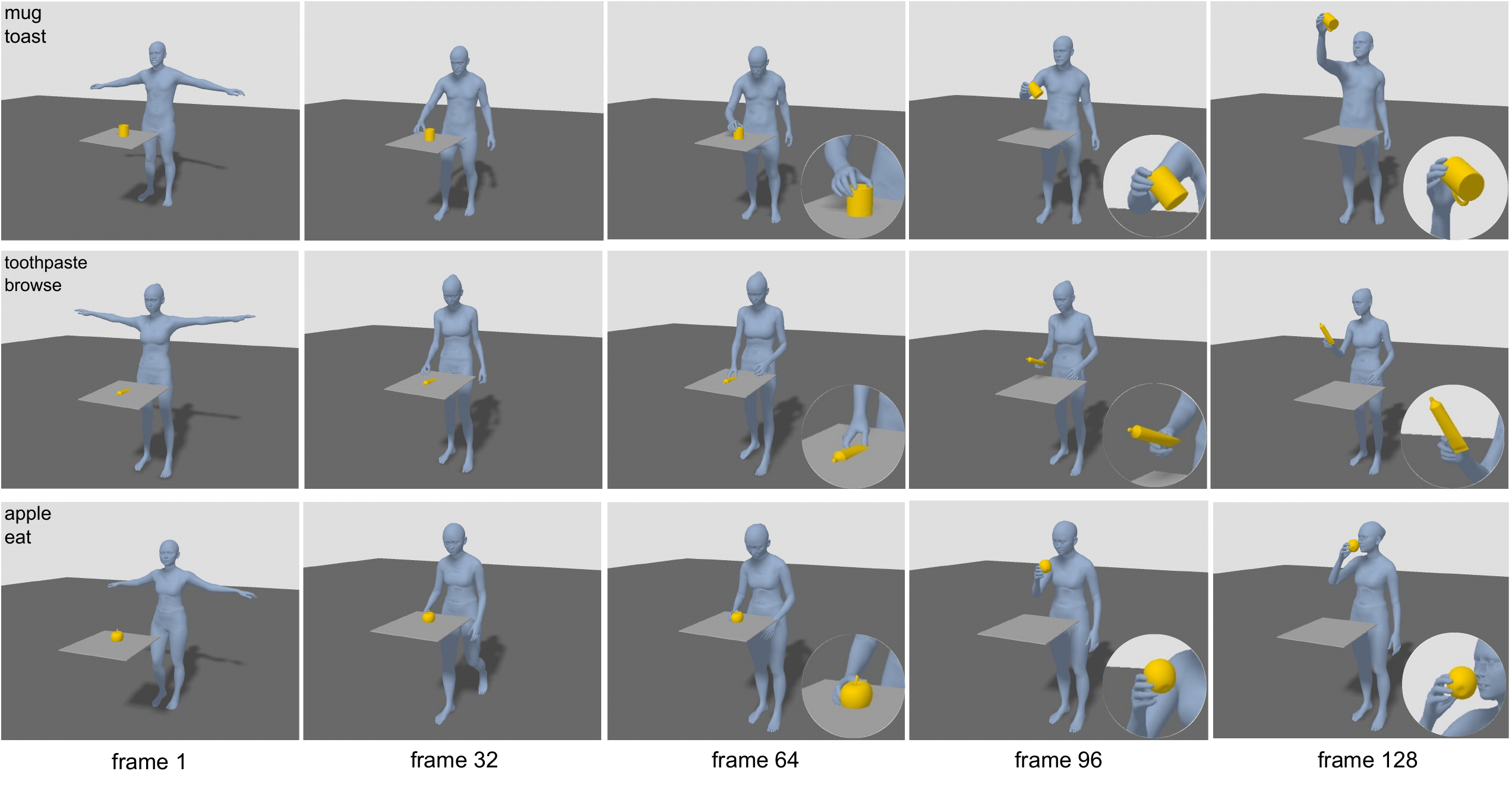}
    \caption{We present generated results of TOHO. Each row represents a human with a distinct shape conducting a task with a novel object unseen during training.}
    \label{fig:full_pipeline}
\end{figure*}
\label{sec:experiments}
\subsection{Object Motion Estimation} 
Synthesizing human-object interactions without object motions is incomplete. Our method generates whole human-object interaction motions with objects moving consistently with humans. Instead of using a parametric model which is computationally expensive and slow, we propose a compact closed-form object motion estimation algorithm that is stable and yields realistic and natural object motion sequences based on human motions in real-time.

As most hand manipulation tasks only involve the right hand (for right-handed people), we focus on the right-hand-object interactions. When the human achieves a stable grasp, the algorithm computes an object motion based on the grasp to finish the task. 
We denote the frame of this stable grasp, i.e., the ‘touching’ pose generated by our pose predictor, as frame 1. Denote the locations of the five markers at five right-hand fingertips as $v^{f_i},i\in\{1, 2, 3, 4, 5\}$ and the marker at the right palm as $v^{f_0}$, $v^{f_i} \in \mathbb{R}^3$. 
At each timestamp $n\in \{2,...,T\}$, we compute the offset vectors of the fingertip markers from the palm as $o_n^{f_i}= v_n^{f_i}- v_n^{f_0}, i\in \{1, 2, 3, 4, 5\}$, and find the rotation $R_n$ that optimally aligns the set of vectors $o_n^{f_{1:5}}$ to $o_1^{f_{1:5}}$ with the Kabsch algorithm, i.e., find $R_n$ s.t.,
\begin{equation}
R_n=\underset{R}{\mathrm{argmin}} \frac{1}{2} \sum^5_{i=1}||o_1^{f_i}-R o_n^{f_i}||_2 \label{rotation}
\end{equation}
Thus, the object orientation at time $n$ is given by:
\begin{equation}
rot_n= R_n^T rot_1, \label{5}
\end{equation}
where $rot_i$ is the rotation matrix of the object orientation at frame $n$. The 6D orientation $r_n$ can be computed from $rot_n$. Then the object translation at frame $n$ is given by:
\begin{equation}
t_n^o=\frac{1}{6}\sum_{i=0}^5 v_n^{f_i}+R^T_n(t_1^o-\frac{1}{6}\sum_{i=0}^5 v_1^{f_i}) \label{6}
\end{equation}
The object motion parameters $(r_{1:T}, t_{1:T}^o)$ are then used to construct the object motion consistent with the right hand.

\section{Experiments}

We train our pipeline on the GRAB dataset \cite{GRAB:2020}, following the same split as \cite{taheri2021goal}. The dataset consists of 51 objects and four manipulation intents: pass, lift, offhand, and use. For the use intent, there are 26 sub-tasks in total. However, some sub-tasks are specific to certain objects and have only one sequence for a subject in the dataset, and some sub-tasks are similar to each other though having different names. Thus, we merge some of the tasks into 6 tasks that have distinct behaviors to demonstrate the effectiveness of our method.
We label the representative frames of each task in the dataset, which requires minimal labor work. For training our pose predictor, we additionally label frames that the human first grasps the object from the table and the following 20 frames as a 'touch' task to indicate the grasping poses.
For the motion inbetweening model, we downsample the framerate of GRAB from 120 fps to 30 fps, slide over each sequence with a skip frame of 16, and chop the sequences into 64 frames as unit training sequences.

\subsection{Qualitative Results}

\noindent\textbf{Full pipeline.} 
A natural human-object interaction motion requires the human to first walk towards the object, grasp it, and then conduct the task. 
Fig. \ref{fig:full_pipeline} shows generated sequences of three humans of \textit{different shapes} performing \textit{distinct tasks} with \textit{unseen objects}.

\noindent\textbf{Upsampling} As our motion inbetweening model generates continuous motions, it is theoretically guaranteed to allow upsampling to arbitrary frames. In the supplementary materials, we show our model generates smooth motion even at 512 frames although trained only with 64 frames sequences.

\begin{figure}
\centering
\begin{subfigure}{0.23\textwidth}
    \includegraphics[width=\textwidth]{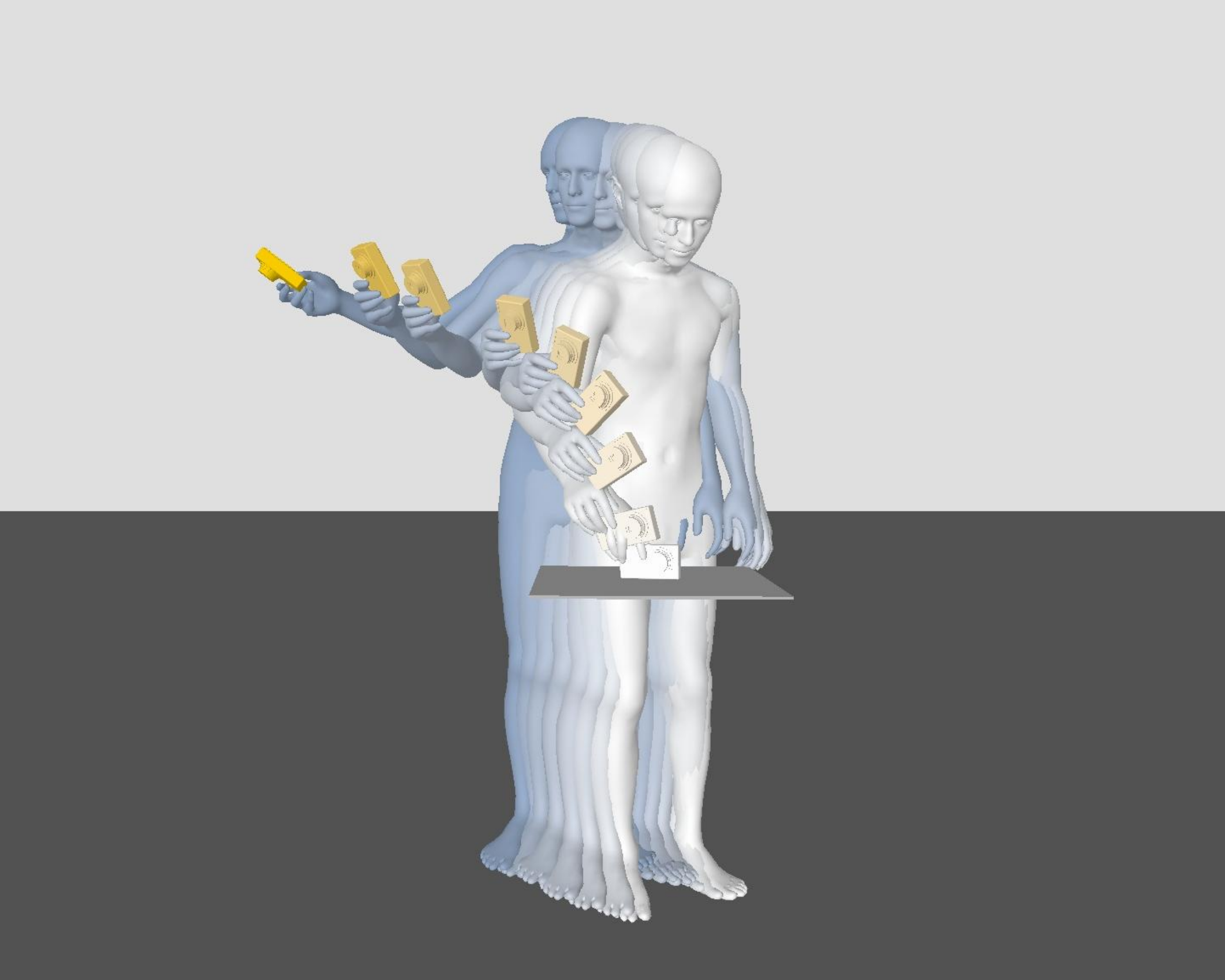}
    \caption{Normal speed}
    \label{fig:vel_a}
\end{subfigure}
\hfill
\begin{subfigure}{0.23\textwidth}
    \includegraphics[width=\textwidth]{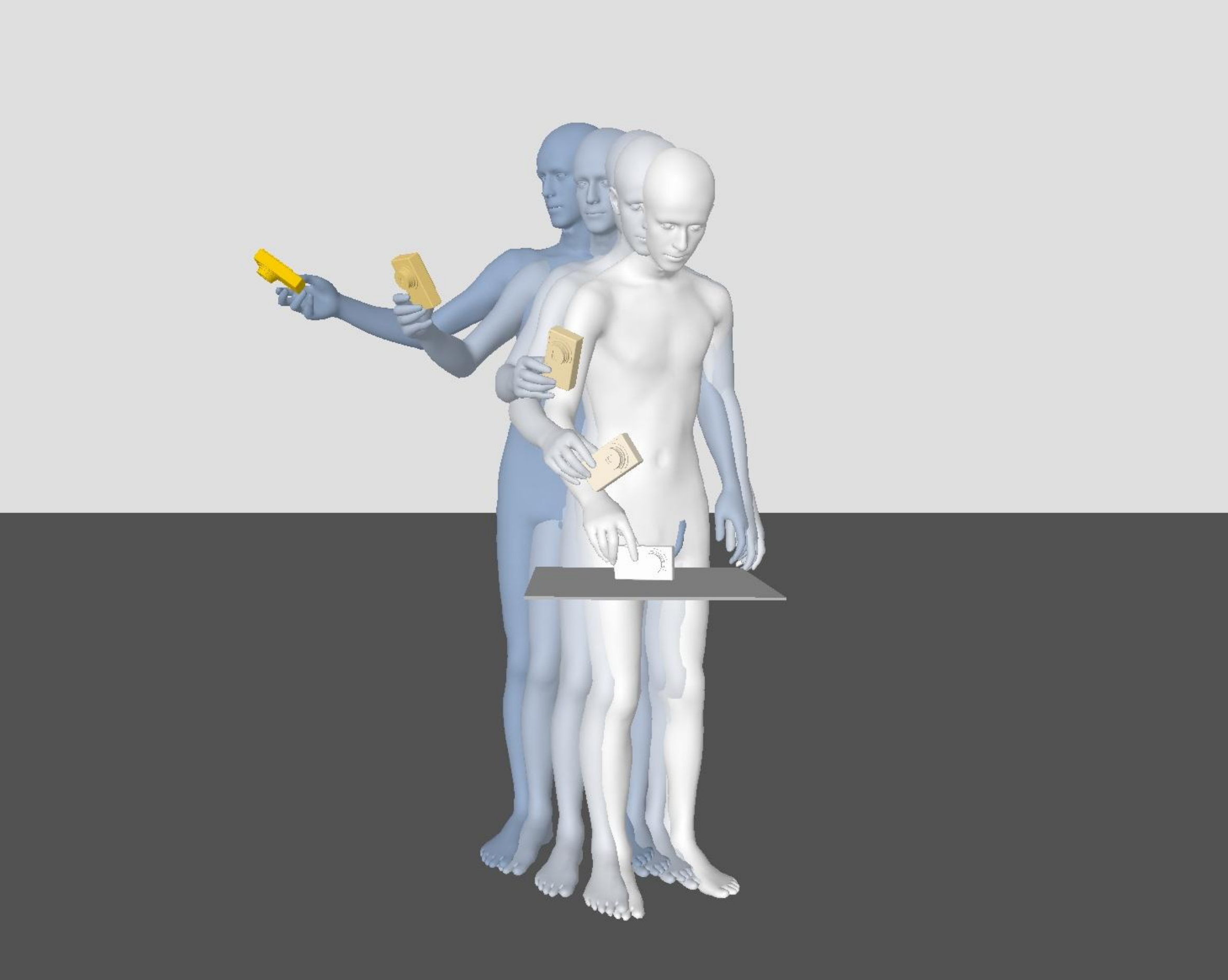}
    \caption{$\times 2$ speed}
    \label{fig:vel_b}
\end{subfigure}
\hfill
\begin{subfigure}{0.23\textwidth}
    \includegraphics[width=\textwidth]{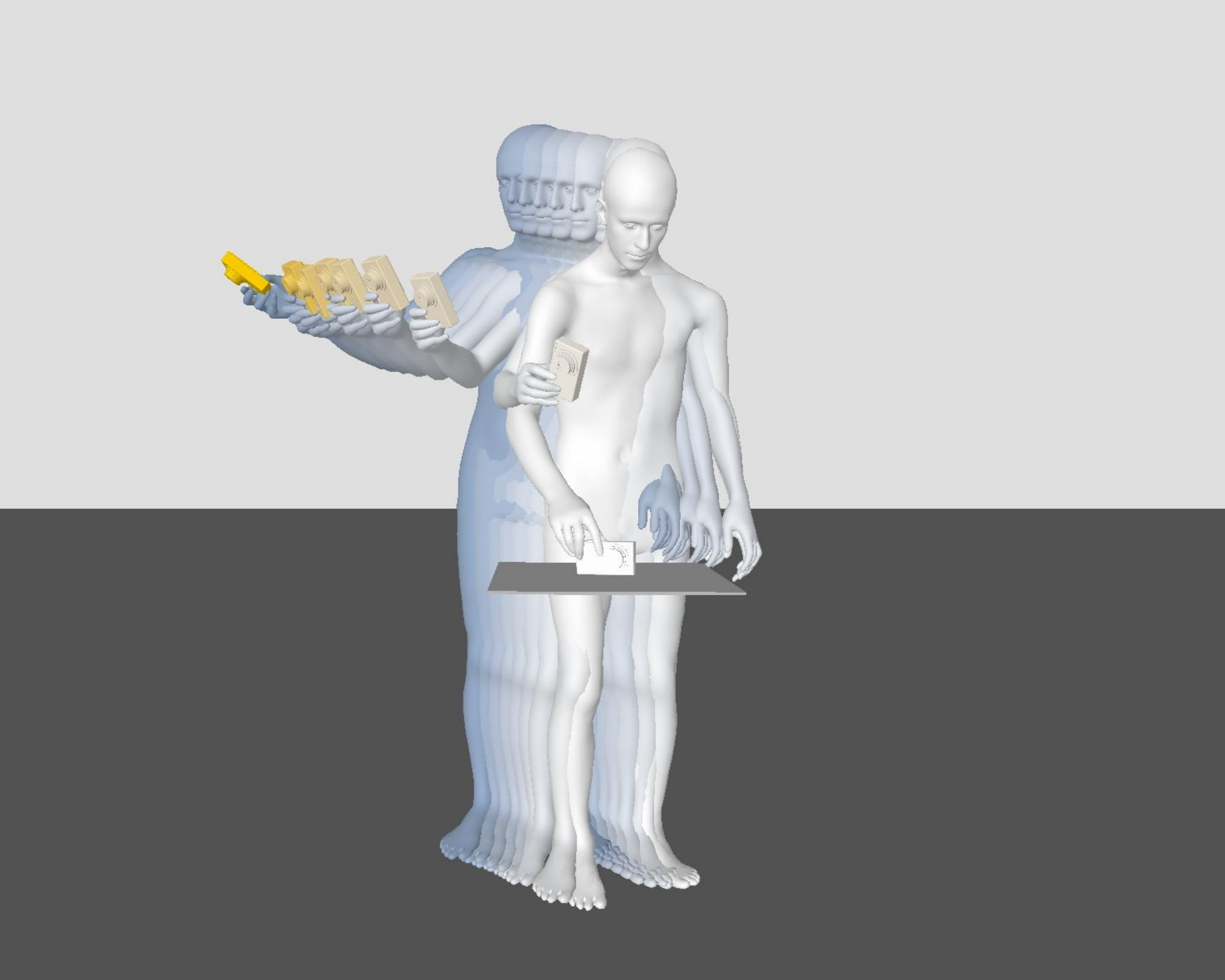}
    \caption{Fast to slow}
    \label{fig:vel_c}
\end{subfigure}
\hfill
\begin{subfigure}{0.23\textwidth}
    \includegraphics[width=\textwidth]{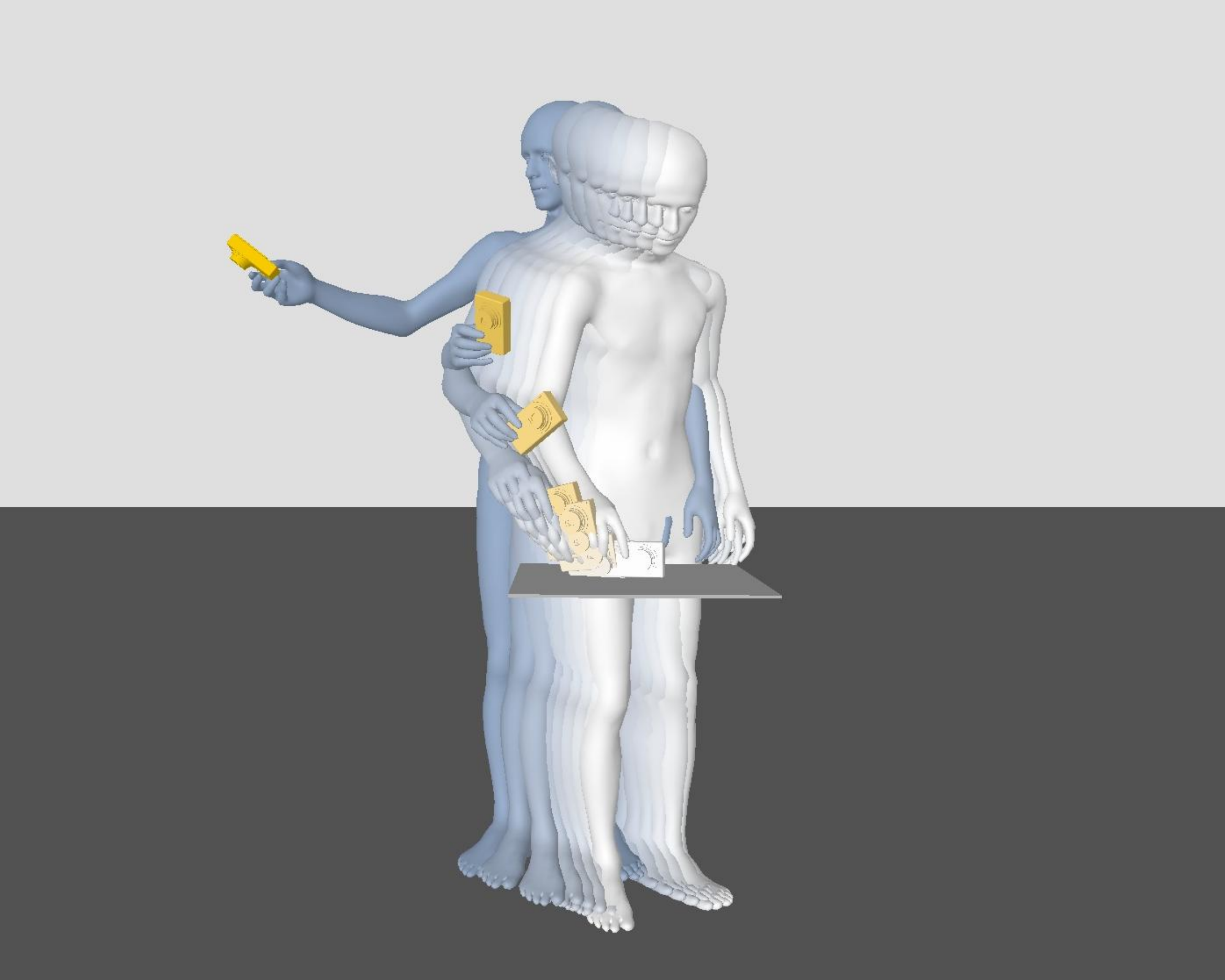}
    \caption{Slow to fast}
    \label{fig:vel_d}
\end{subfigure}
\caption{Examples of motion velocity adjustment by modifying the temporal coordinate $\tau$. a) A 64-frame generated result of normal speed. b) Speed up the sequence by uniformly sampling two times fewer values from $\tau \in [0, 1]$, which gives a result of doubled velocities. c) The human swiftly lifts the object and then slowly passes it, which is done by sampling sparsely near $0$ and densely near $1$. d) The human slowly lifts the object and swiftly passes it, which uses a reverse sampling scheme of c).}
\label{figure:velocity}
\end{figure}

\noindent\textbf{Velocity Modification} Another advantage of our method is that by designing the temporal coordinate $\tau$, users can generate motions with different velocities at specific parts of the sequence with the same model weights. This allows for affluent post-artworks to output motions reflecting divergent states of the human. Fig. \ref{figure:velocity} shows the motions generated with the same inferenced model weights but different temporal inputs. The four inputs include a uniform sampling of $\tau$ on $[0, 1]$, a uniform sampling that is two times sparser, a non-uniform sampling that is denser at the end, and a sampling denser at the start. Here, we design the lengths of intervals of the non-uniform sampling as geometric sequences.

\subsection{Quantitative Results}
\subsubsection{Motion Diversity}
We calculate the Average L2 Pairwise Distance (APD) \cite{yuan2020dlow} of the object translation offsets predicted by our object position estimator, given the same objects and task types but different human shapes, to be 0.19 ($m$). This indicates our model generates diverse ending object positions for different human shapes within a reasonable range. We report the APD of our full generated motions on GRAB to be 0.34.

\subsubsection{Motion Inbetweening}

\begin{table}[t]
\centering
\caption{
Comparisons of motion-infilling performance. Second row shows results of pose prediction using ground truth trajectories and third row shows results of prediction of both the poses and trajectories. Boldface represents best results.
}
\resizebox{\columnwidth}{!}{
\begin{tabular}{clcccc}
\toprule
 & \begin{tabular}{l}
     Methods
 \end{tabular} & \begin{tabular}{@{}c@{}}ADE \\ ($\downarrow$)\end{tabular} &
 \begin{tabular}{@{}c@{}}Skat \\ ($\downarrow$)\end{tabular} & 
\multicolumn{2}{c}{\begin{tabular}{ll} \multicolumn{2}{c}{PSKL-J ($\downarrow$)} \\ \midrule (P, GT) &
(GT, P) \end{tabular}}  \\
\midrule
\makecell{GT Traj + local \\ motion \\ infilling} & 
\begin{tabular}{l}
CNN-AE \cite{kaufmann2020convolutional}\\
LEMO \cite{zhang2021priors} \\
PoseNet \cite{wang2021synthesizing} \\
SAGA-Local \cite{wu2022saga}\end{tabular} &
\begin{tabular}{@{}c@{}}
0.091\\
0.083 \\
0.090 \\
\textbf{0.079}\end{tabular} & 
\begin{tabular}{@{}c@{}}
0.245\\
0.152 \\
0.236 \\
\textbf{0.137}\end{tabular} & 
\multicolumn{2}{l}{\begin{tabular}{ll}
\begin{tabular}{l}
0.804\\
0.507 \\
0.611 \\
\textbf{0.377}\end{tabular} &
\begin{tabular}{l}
0.739\\
0.447 \\
0.668 \\
\textbf{0.372}\end{tabular} \end{tabular}}
\\
\midrule
\makecell{Traj + local \\ motion \\ infilling} &
\begin{tabular}{l}
Route+PoseNet \cite{wang2021synthesizing}\\
SAGA \cite{wu2022saga} \\
\textbf{Ours}\end{tabular} &
\begin{tabular}{@{}c@{}}
0.219\\
\textbf{0.083} \\
0.113\end{tabular} &
\begin{tabular}{@{}c@{}}
0.575\\
0.394 \\
\textbf{0.247}\end{tabular} &
\multicolumn{2}{l}{\begin{tabular}{ll}
\begin{tabular}{l}
0.955\\
0.772 \\
\textbf{0.232}\end{tabular} &
\begin{tabular}{l}
0.884\\
0.609 \\
\textbf{0.218}\end{tabular}  \end{tabular}} \\
\midrule
\makecell{GRAB$\ast$} &
\begin{tabular}{l}
IMoS \cite{ghosh2023imos}\\
\textbf{Ours}\end{tabular} &
\begin{tabular}{@{}c@{}}
0.087\\
0.081\end{tabular} &
\begin{tabular}{@{}c@{}}
0.413\\
0.278\end{tabular} &
\multicolumn{2}{l}{\begin{tabular}{ll}
\begin{tabular}{l}
0.411\\
0.231\end{tabular} &
\begin{tabular}{l}
0.339\\
0.219\end{tabular}  \end{tabular}} \\
\bottomrule
\end{tabular}
}
{\tiny $\ast$ Tested on the GRAB dataset with sequences starting from grasping poses, following the setting of IMoS.}
\vspace{-0.05in}
\label{table:motion}
\end{table}

We conduct experiments to demonstrate the effectiveness of our motion inbetweening model. To our knowledge, our model is the first generalizable motion inbetweening model that generates continuous motions.

\noindent\textbf{Baselines.} SAGA \cite{wu2022saga} proposed a CNN-based motion-infilling network that predicts both the human poses and their root translations. Wang et al. \cite{wang2021synthesizing} proposed an LSTM-based model, namely Route+PoseNet, to infill motion sequences. We take these two as our baselines as they are the closest to our settings. We also compare our model with some existing works, including the convolution autoencoder network (CNN-AE) \cite{kaufmann2020convolutional}, LEMO \cite{zhang2021priors}, and PoseNet \cite{wang2021synthesizing}, that take the ground truth trajectory as inputs and infill the local motions. We use the same body markers as in \cite{wu2022saga} for fair comparisons.

\noindent\textbf{Evaluation Metrics.} We use the same evaluation metrics in \cite{wu2022saga}. 1) \textit{3D marker accuracy.} We compute the Average L2 Distance (ADE) between the reconstructed marker sequences and the ground truth. 2) \textit{Foot skating.} We follow \cite{yuan2020dlow} to decide on skating frames and report their ratio in the full sequences. 3) \textit{Motion smoothness.} We measure the smoothness of the generated sequences by computing the Power Spectrum KL Divergence of their joints (PSKL-J) \cite{zhang2021priors} to compare with the ground truth. We report the scores of PSKL-J in both directions as they are asymmetric. Note a lower PSKL-J score represents a closer generated distribution to the ground truth.

\noindent\textbf{Results.} In Table \ref{table:motion}, we show the results of our method compared to the previous works mentioned above. We experiment on both GRAB \cite{GRAB:2020} and AMASS \cite{AMASS:ICCV:2019} datasets following the settings of \cite{wu2022saga} to test our motion inbetweening model. We show the results of using the ground truth trajectory for \cite{kaufmann2020convolutional, zhang2021priors} as they only infill local motions. We also show the results of PoseNet and SAGA fed with ground truth trajectory to predict local poses. Then, we evaluate both the predicted pose and trajectory of our method against SAGA and Route+PoseNet. The results show that our model gives better scores of PSKL-J in both directions than previous works, which indicates that the continuity of our model facilitates the generation of smoother sequences. Our method also yields lower skating effects than SAGA and Route+PoseNet and achieves a comparable ADE score to SAGA. The results show that our method is on par with SOTA methods while providing the benefits of upsampling and velocity adjustment through its continuity property.

We report the results of IMoS-generated sequences to give a sense of how our method is compared with IMoS \cite{ghosh2023imos}. As IMoS is trained only on the GRAB dataset with sequences starting from the grasping poses, we also report our results tested in the same setting. We uniformly sample 60 frames from TOHO as motion sequences and interpolate the 15 generated frames of IMoS into 60-frame motion clips, such that all motion clips have the same length. Note this is not an exact comparison as IMoS is an autoregressive model while ours is a motion-infilling model. Sequences starting from the grasping poses tend to have small root translation movements, which leads to a relatively higher skating ratio.

\subsubsection{Human-Object Motion}

\begin{table}[t]
\centering
\caption{
Comparisons of our generated human-object interaction sequences to the ground truth. We report the results of the entire motion, and of the grasping poses generated by our pose predictor.
}
\resizebox{\columnwidth}{!}{
\begin{tabular}{ccccccc}
\toprule
&
\multicolumn{3}{c}{\begin{tabular}{ccc} \multicolumn{3}{c}{Contact Ratio ($\uparrow$)} \\ \midrule Max. & Min. & Avg. \end{tabular}} &
\multicolumn{3}{c}{\begin{tabular}{ccc} \multicolumn{3}{c}{Interp. Depth (m)} \\ \midrule Max. & Min. & Avg. \end{tabular}}
\\
\midrule
GT & 
\multicolumn{3}{c}{\begin{tabular}{ccc} 1.00 & 0.99 & 1.00 \end{tabular}} &
\multicolumn{3}{c}{\begin{tabular}{ccc} 0.007 & 0.005 & 0.006 \end{tabular}}
\\
Ours & 
\multicolumn{3}{c}{\begin{tabular}{ccc} 0.93 & 0.77 & 0.85 \end{tabular}} &
\multicolumn{3}{c}{\begin{tabular}{ccc} 0.012 & 0.006 & 0.007 \end{tabular}}
\\
\midrule
Pose Predictor &
\multicolumn{3}{c}{\begin{tabular}{ccc} \multicolumn{3}{c}{0.91} \end{tabular}} &
\multicolumn{3}{c}{\begin{tabular}{ccc} \multicolumn{3}{c}{0.008} \end{tabular}}
\\
\bottomrule
\end{tabular}
}
\vspace{-0.2in}
\label{table:object}
\end{table}

We conduct experiments to evaluate the realisticness of our human-object motions by computing the hand-object contact ratio \cite{zhang2020withoutpeople} and the largest hand-object interpenetration depth along the whole motion sequence. We report the maximum, minimum, and average contact ratio and interpenetration depth along object movements. Our object motion estimation algorithm captures the stable grasp at the initial grasping frame and keeps this hand-object relationship across the sequence, so we also report the contact ratio and interpenetration depth of our pose predictor generated results. 
Table \ref{table:object} shows that our method well preserves good grasps during the human-object interaction and yields results that are comparable to the ground truth.

\subsubsection{Ablation Study}

\textbf{(1) Human shape for object parameters sampler.} To evaluate the effects of human shapes in estimating final object positions, we conduct experiments by training an object parameters sampler which does not take the human shape as input. We report the average distance from the predicted object positions to the ground truth of this model as 0.073 (m), while ours taking the human shape information as input during training has a result of 0.048 (m). The results suggest that human shape is an impactful factor in estimating the final object position given a task type. \textbf{(2) Losses of the motion inbetweening model.} In Table \ref{table:motion_ablation}, we show the results of our motion inbetweening model trained without the foot-ground contact loss/surface marker loss. All three models are trained and tested on the GRAB dataset. \textbf{(3) Object motion estimation.} To demonstrate the effectiveness of aligning the orientation of the object to the fingertip rotations, we compare our method with one without $R_n$, i.e., simply averaging the fingertip trajectories as the object trajectory and keeping the object orientation still. We show the results in Table \ref{table:object_ablation}. For both methods, we use the ground truth human motions to compute the object trajectories.

\begin{table}[t]
\centering
\caption{
Ablation study of our motion inbetweening model. $\mathcal{L}_C$ is the foot-ground contact loss. $\mathcal{L}_v$ is the surface marker loss. Boldface represents best results.
}
\resizebox{\columnwidth}{!}{
\begin{tabular}{lcccc}
\toprule
  & \begin{tabular}{@{}c@{}}ADE \\ ($\downarrow$)\end{tabular} &
 \begin{tabular}{@{}c@{}}Skat \\ ($\downarrow$)\end{tabular} & 
\multicolumn{2}{c}{\begin{tabular}{ll} \multicolumn{2}{c}{PSKL-J ($\downarrow$)} \\ \midrule (P, GT) &
(GT, P) \end{tabular}} 
\\
\midrule
\begin{tabular}{l}
Ours-w/o $\mathcal{L}_C$\\
Ours-w/o $\mathcal{L}_v$\\
Ours\end{tabular} &
\begin{tabular}{@{}c@{}}
0.088\\
0.096 \\
\textbf{0.079}\end{tabular} &
\begin{tabular}{@{}c@{}}
0.229\\
0.201 \\
\textbf{0.177}\end{tabular} &
\multicolumn{2}{l}{\begin{tabular}{ll}
\begin{tabular}{l}
0.242\\
0.264 \\
\textbf{0.219}\end{tabular} &
\begin{tabular}{l}
0.235\\
0.256 \\
\textbf{0.208}\end{tabular}  \end{tabular}} \\
\bottomrule
\end{tabular}
}
\vspace{-0.05in}
\label{table:motion_ablation}
\end{table}

\begin{table}[t]
\centering
\caption{
Ablation study of our object motion estimation algorithm. $R_n$ is the optimal alignment rotation between the current and first frames of hand vectors.
}
\resizebox{\columnwidth}{!}{
\begin{tabular}{lcccccc}
\toprule
&
\multicolumn{3}{c}{\begin{tabular}{ccc} \multicolumn{3}{c}{Contact Ratio ($\uparrow$)} \\ \midrule Max. & Min. & Avg. \end{tabular}} &
\multicolumn{3}{c}{\begin{tabular}{ccc} \multicolumn{3}{c}{Interp. Depth (m)} \\ \midrule Max. & Min. & Avg. \end{tabular}}
\\
\midrule
Ours-w/o$R_n$ & 
\multicolumn{3}{c}{\begin{tabular}{ccc} 0.88 & 0.42 & 0.66 \end{tabular}} &
\multicolumn{3}{c}{\begin{tabular}{ccc} 0.019 & 0.008 & 0.013 \end{tabular}}
\\
Ours & 
\multicolumn{3}{c}{\begin{tabular}{ccc} \textbf{0.98} & \textbf{0.89} & \textbf{0.93} \end{tabular}} &
\multicolumn{3}{c}{\begin{tabular}{ccc} \textbf{0.009} & \textbf{0.004} & \textbf{0.007} \end{tabular}}
\\
\bottomrule
\end{tabular}
}
\vspace{-0.15in}
\label{table:object_ablation}
\end{table}

\section{Conclusion}
\label{sec:conclusion}
In this paper, we introduce TOHO, the first approach to synthesizing continuous task-oriented human-object interaction motions with unseen objects. TOHO generates complete human-object motions to conduct specific tasks with the task type, the object's initial information, and the starting human status as the only inputs. We address the synthesis process in four steps: 1) we estimate the ending object position given the task type; 2) use object positions and task labels to estimate keyframe poses; 3) the motion inbetweening model then generates continuous motions to infill the keyframes; 4) apply a novel closed-form object motion estimation algorithm to produce an object motion consistent to the human motion. The evaluation results show our framework can generate natural and realistic motions, and our generated motions are continuous, allowing for arbitrary upsampling and velocity adjustment. Future work may include generating interaction motions in scenes with occlusions and synthesizing physically corrected motions.

\noindent \textbf{Acknowledgment.} This research is supported by the National Research Foundations, Singapore under its AI Singapore Programme (AISG Award No: AISG2-PhD-2022-01-032[T]). It is also supported under the RIE2020 Industry Alignment Fund Industry Collaboration Projects (IAF-ICP) Funding Initiative, as well as cash and in-kind contribution from the industry partner(s). It is partially supported by Singapore MOE AcRF Tier 2 (MOE-T2EP20221-0011).

{\small
\bibliographystyle{ieee_fullname}
\bibliography{bib}
}

\end{document}